\documentclass{article}

\usepackage{microtype}
\usepackage{graphicx}
\usepackage{booktabs} 
\usepackage{subcaption}

\usepackage{amsmath,amsfonts,bm}

\def\eqref#1{equation~\ref{#1}}

\def\1{\bm{1}}

\def\vtheta{{\bm{\theta}}}

\def\vx{{\bm{x}}}

\DeclareMathAlphabet{\mathsfit}{\encodingdefault}{\sfdefault}{m}{sl}
\SetMathAlphabet{\mathsfit}{bold}{\encodingdefault}{\sfdefault}{bx}{n}

\DeclareMathOperator*{\argmax}{arg\,max}

\usepackage{hyperref}

\usepackage[accepted]{icml2025}

\usepackage{amsmath}
\usepackage{amssymb}
\usepackage{mathtools}
\usepackage{amsthm}
\usepackage{multirow} 

\usepackage[capitalize,noabbrev]{cleveref}

\theoremstyle{plain}

\theoremstyle{definition}

\theoremstyle{remark}

\usepackage[textsize=tiny]{todonotes}

\icmltitlerunning{Submission and Formatting Instructions for ICML 2025}

\begin{document}

\twocolumn[
\icmltitle{FairDropout: Using Example-Tied Dropout to Enhance Generalization of Minority Groups}



\icmlsetsymbol{equal}{*}

\begin{icmlauthorlist}
\icmlauthor{Geraldin Nanfack}{yyy,comp}
\icmlauthor{Eugene Belilovsky}{yyy,comp}
\end{icmlauthorlist}

\icmlaffiliation{yyy}{Concordia University}
\icmlaffiliation{comp}{Mila – Quebec AI Institute}

\icmlcorrespondingauthor{Geraldin Nanfack}{geraldin.nanfack@concordia.ca}

\icmlkeywords{Machine Learning, ICML}

\vskip 0.3in
]



\printAffiliationsAndNotice{}

\begin{abstract}
Deep learning models frequently exploit spurious features in training data to achieve low training error, often resulting in poor generalization when faced with shifted testing distributions. To address this issue, various methods from imbalanced learning, representation learning, and classifier recalibration have been proposed to enhance the robustness of deep neural networks against spurious correlations. In this paper, we observe that models trained with empirical risk minimization tend to generalize well for examples from the majority groups while memorizing instances from minority groups.
Building on recent findings that show memorization can be localized to a limited number of neurons, we apply example-tied dropout as a method we term \textit{FairDropout}, aimed at redirecting this memorization to specific neurons that we subsequently drop out during inference. We empirically evaluate FairDropout using the subpopulation benchmark suite encompassing vision, language, and healthcare tasks, demonstrating that it significantly reduces reliance on spurious correlations, and outperforms state-of-the-art methods.
\end{abstract}

\section{Introduction}\label{sec:intro}

Deep neural networks trained with empirical risk minimization (ERM) continue to achieve remarkable performance across various tasks. However, these ERM-trained models often experience a drop in predictive performance when facing a variety of subpopulation shifts~\citep{yang2023change,QuioneroCandela2009Dataset}. In particular, if datasets contain spurious features, i.e., patterns that are highly predictive of the training labels but not causally related to the target, ERM may fail to learn robust features that generalize well across subpopulation shifts~\citep{geirhos2020shortcut}. ERM-trained models may thus rely on non-robust features, such as image backgrounds or hair colors, for predictions. However, in certain applications, reliance on such spurious features (e.g., hair colors) can hurt the \textit{fairness}, raising potential safety concerns in deployment~\citep{amodei2016concrete}. 

Several works have been proposed to tackle the challenge of learning more robust features in the presence of spurious correlations. A common practical setup assumes a known group partition within datasets~\citep{liu2021just,sagawa2019distributionally}. In this context, training labels and spurious features are often highly correlated in a specific group of training data, but this correlation does not hold in testing data. Consequently, models trained naively can easily maximize training performance by relying on spurious features, yet suffer a significant drop in worst-group performance during testing when the spurious correlation fails to hold. 
Most of the existing work in this area assumes the presence of group annotations in the training set to learn more robust-to-spurious-correlation features.
For instance, GroupDRO~\citep{sagawa2019distributionally} directly minimizes the worst group error. However, such approaches depend heavily on prior knowledge of group labels in training data, which can be impractical in large-scale settings. There exist other methods that do not assume this availability of group annotations on the training set. For example, \citet{kirichenko2022last} observed that ERM captures both \textit{core} (or robust) and spurious features. They then proposed a two-stage approach, known as DFR, in which the first stage involves ERM training, and the second stage re-trains the classifier using a group-balanced validation dataset. While DFR has effectively improved worst-group performance, it still relies on group annotations in the validation set to create the balanced set necessary for down-weighting spurious features. 

In this work, we hypothesize that reducing example-level memorization can reduce the reliance on spurious correlations without the need for group labels.
Building on recent insights connecting memorization with generalization~\citep{baldock2021deep,maini2023can}, we investigate this connection in the context of spurious correlations for the first time. We apply a recent technique for localizing memorization, initially developed for label noise in small networks, and scale it to larger networks in the spurious correlation setting. Our method, termed \textit{FairDropout}, fairly distributes \textit{memorizing neurons} during training and selectively drops out these neurons during inference. Our contributions are threefold: (i) We identify a discrepancy in generalization behavior between majority and minority groups, linking this difference to memorization effects. (ii) We demonstrate that example-tied dropout, previously used in label noise settings with smaller networks~\citep{maini2023can}, can be scaled and applied to larger architectures (e.g., BERT~\citep{sung2019pre}) in the context of spurious correlation, under the name \textit{FairDropout}. (iii) We evaluate FairDropout on the subpopulation benchmark suite, showing significant improvements in worst-group accuracy across image, medical (X-Ray), and language tasks.

\section{Related Work}\label{sec:related_work}
Methods have been proposed to fight against spurious correlation. 

\subsection{Using Training Group Information} Most methods that fight against spurious correlation assume to have training group annotations. Some approaches directly modify ERM; for example, groupDRO~\citep{sagawa2019distributionally} and its variant CVaRDRO~\citep{duchi2021learning} aim to minimize the worst group error rather than the average error used by ERM. Similarly, when group information is available, methods from out-of-distribution generalization~\citep{arjovsky2019invariant,krueger2021out,wald2021calibration,krueger2021out} can be framed to learn more robust-to-spurious-correlation features. Other approaches use training group information to synthetically augment minority group samples through generative modeling \citep{goel2020model}. Reweighting and subsampling techniques can also be used to balance majority and minority groups~\citep{sagawa2020investigation,byrd2019effect}. However, all these works share a major limitation:  they rely on group information, which is difficult to obtain for large datasets. Indeed, manually annotating group labels requires task-specific expertise, making it prohibitively costly.

\subsection{Without Using Training Group Information} Given the expensive cost of manual group annotation, there has been a growing interest in combating spurious correlation without group annotations in the training set. Some methods observe the training dynamics of SGD and introduce margin-based regularization terms to promote more robust feature learning~\citep{pezeshki2021gradient, puli2023don}. Among the most popular approaches are two-stage algorithms that do not assume prior group annotation for the training set; these methods typically begin with ERM to infer minority group samples. In the second stage, robustness to spurious correlations is enhanced, for instance through contrastive learning~\citep{zhang2022correct} or by up-weighting the inferred minority group samples~\citep{qiu2023simple, liu2021just}. A recent study~\citep{kirichenko2022last} underscores the importance of the first stage ERM, by showing that ERM learns both spurious and \textit{core} (or robust-to -spurious-correlation) features. It then proposes retraining only the classifier head in a second stage using a group-balanced validation set. This approach has been extended to HTT-DFR~\citep{hameed2024not}, where the second phase retrains a sparse network. Inspired by ERM’s ability to capture core features, our work aims to reduce the reliance on spurious correlations by focusing on memorizing neurons, offering a simpler alternative to computationally demanding two-stage methods.

\subsection{Memorization and Generalization Links} Recent research has made strides in understanding the relationship between generalization and memorization in deep learning. Memorization is often defined as the model's capacity to correctly predict \textit{atypical} examples with potentially wrong patterns~\citep{maini2023can}. Several studies, such as~\citet{jiang2021characterizing, carlini2019distribution}, have introduced metrics to quantify the degree to which an example is regular or atypical. Some initial works observed that memorization primarily occurs in later layers of the network~\citep{baldock2021deep, stephenson2021geometry}, while recent findings from~\citet{maini2023can} suggest that memorization can emerge at any depth within a network. Additionally, \citet{maini2023can} introduces a method for localizing memorization by determining the minimum number of neurons required to change a model’s predictions. Our work builds on this method, applying it beyond the context of label noise, as in the original study, to address spurious correlations. Various studies on mechanistic interpretability further reveal that certain neurons in deep networks specialize in capturing specific patterns, often associated with spurious features~\citep{zenke2017continual,cheung2019superposition,hendel2023context,kalibhat2023identifying}.
\section{Methods}
\label{sec:method}
This section begins by outlining the problem, followed by an analysis of minority group example memorization, and then introduces FairDropout.

\begin{figure*}[!t]
\centering
\begin{subfigure}[]{0.45\linewidth}
\includegraphics[width=\textwidth]{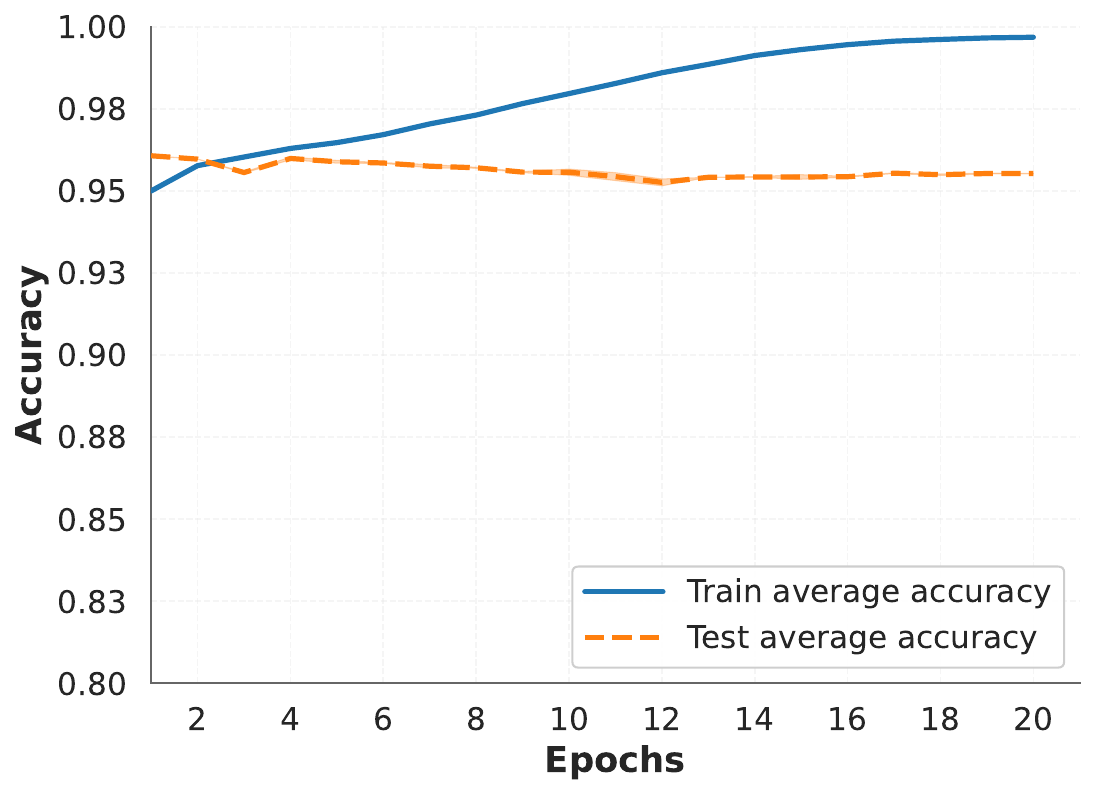}
\end{subfigure}
\begin{subfigure}[]{0.45\linewidth}
\includegraphics[width=\textwidth]{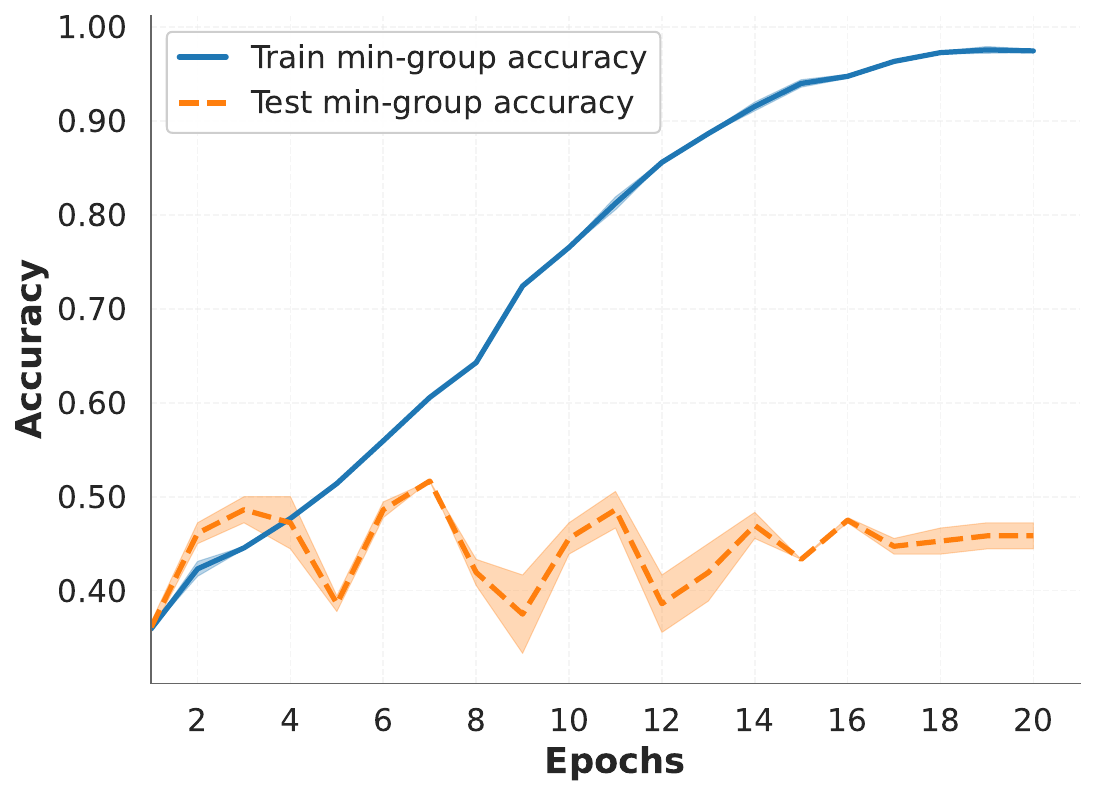}
\end{subfigure}
\caption{Discrepancy in generalization behaviors between majority and minority groups on CelebA. Left: average train and test accuracy are plotted. Right: minimum-group train and test accuracy are plotted. We observe that models trained exhibit a large generalization gap on minority groups, a synonym of minority-group overfitting.}
\label{fig:minority_majority_groups}
\end{figure*}

\subsection{Problem Description}\label{sec:problem_desc}
We consider a classification problem with a training sample $\mathcal{D_{\text{tr}}}=\{(\vx,y)\}_{i=1}^{N}$ drawn from a training distribution $p_{\text{tr}}$, where $\vx_i\in \mathcal{X}$ is the input and $y_i\in \mathcal{Y}$ is its class label. We further assume the existence of a spurious attribute $a \in \mathcal{A}$, which is non-predictive of $y$~\citep{ye2024spurious}. We denote by groups, the pairs  $g:= (y,a) \in \mathcal{Y}\times \mathcal{A}:= \mathcal{G}$. Since $a$ is not predictive in $y$, a correlation between $y$ and $a$ in the training distribution $p_\text{tr}$, but not in the test distribution $p_{te}$ can cause models trained on $\mathcal{D_{\text{tr}}}$ to perform poorly on groups where this correlation does not hold. 

For example, CelebA~\citep{liu2015deep} is one of the most popular datasets in the spurious correlation literature. The common task is to predict the hair color from celebrity faces ($\mathcal{Y}=\{\text{blond hair}, \text{non-blond hair}\}$), gender serves as the spurious attribute ($a\in \{\text{woman}, \text{man}\}$). In the CelebA training set, only $1\%$ of faces are from the group of blond men. Consequently, trained models may rely on the spurious gender feature to predict hair color, and average test performance alone may not fully capture model robustness. Instead, evaluating performance based on worst-group accuracy, such as accuracy on blond men, becomes critical for assessing model robustness.

Formally, given a parameterized model $f_\vtheta: \mathcal{X}\longrightarrow \mathcal{Y} $, the goal of learning in the presence of spurious correlation is to find the model that will minimize the worst-group expected error 
\begin{equation}
    \max_{g\in \mathcal{G}} \quad \mathbb{E} \left[\ell_{0-1}\left(f_\vtheta\left(\vx\right), y\right) | g\right],
\end{equation}
where $\ell_{0-1}\left(f_\vtheta(\vx), y\right) = \mathbf{1}\left[f_\vtheta(\vx)\not= y\right]$ is the 0-1 loss~\citep{liu2021just}. We focus on the scenario where group information is unavailable in the training set but accessible at test time for evaluation purposes.  

\subsection{Memorization of Minority Group Examples}
\label{sec:memorization}

We begin our analyses with ERM on CelebA, which is the most popular, real-world, and large dataset frequently used to study spurious correlations, providing insights that generalize to real-world settings. As previously described, CelebA comprises four groups: $\{\text{blond hair}, \text{non-blond hair}\}\times \{\text{woman}, \text{man}\}$, with the minority group being (blond, man). We train ResNet-50~\citep{he2016deep} using ERM and monitor train/test performance across these different groups.

\begin{figure*}[!t]
\centering
\begin{subfigure}[]{0.45\linewidth}
\includegraphics[width=\textwidth]{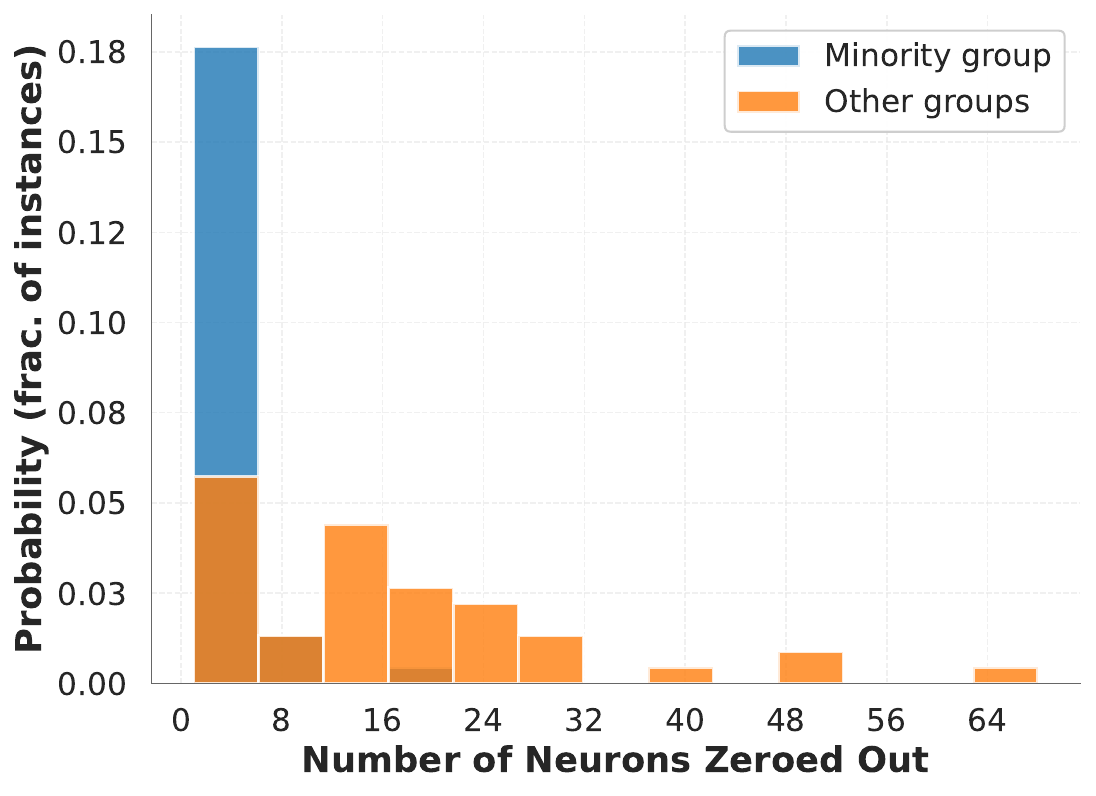}
\caption{Number of Neurons to Flip Predictions.}
\label{fig:number_neurons}
\end{subfigure}
\begin{subfigure}[]{0.45\linewidth}
\includegraphics[width=\textwidth]{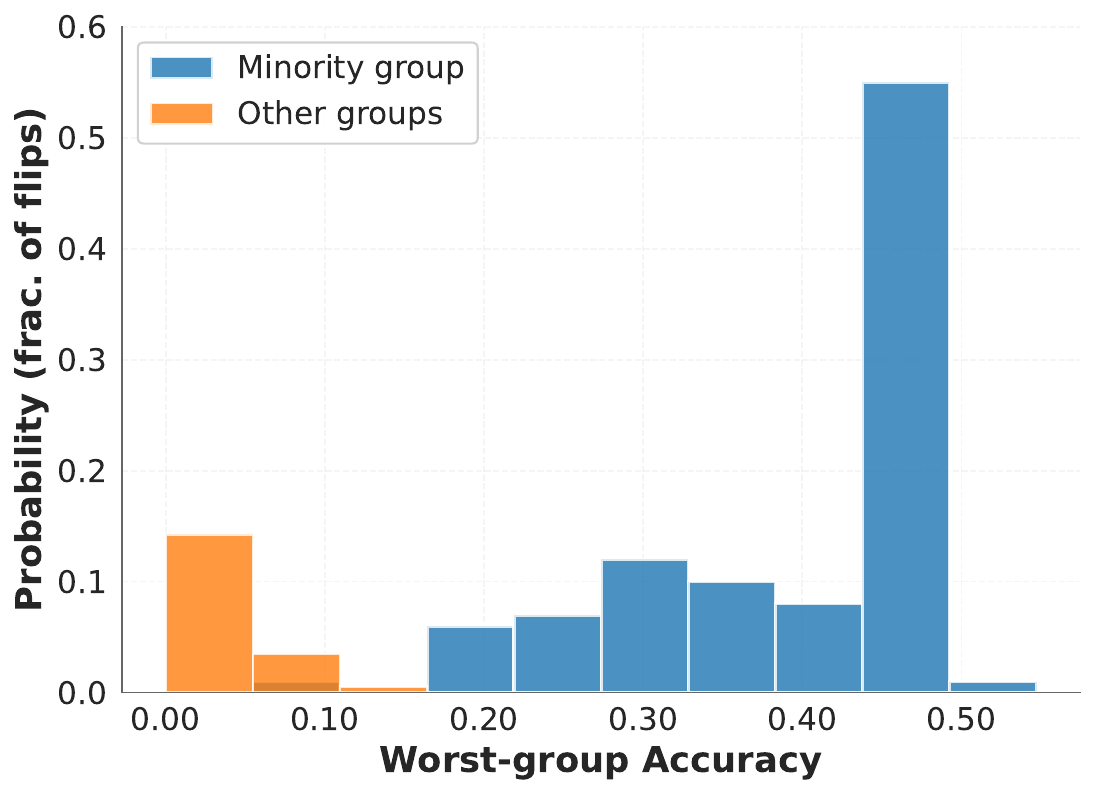}
\caption{Impact on training Worst-group Accuracy.}
\label{fig:impact_worst_group_accuracy}
\end{subfigure}
\caption{
For each example in a random subset of 100 samples from the minority group and 100 from other groups, we iteratively remove the most critical neurons from
a ResNet-50 model trained on the CelebA dataset, until the example’s prediction flips. (a) We observe that minority-group examples require fewer neurons to flip their prediction. (b) After dropping the most critical neurons from examples in different groups, we report the worst-group accuracy on the training set. We find that the worst-group accuracy is consistently less affected when critical neurons from minority-group examples are dropped, compared to those from other groups. This suggests that minority-group examples are being memorized.}
\label{fig:analyze_memorization}
\end{figure*}

Fig.~\ref{fig:minority_majority_groups} shows both average and worst group performance. It can be observed in Fig.~\ref{fig:minority_majority_groups}, a discrepancy in generalization behaviors between the majority groups (represented by the average performance, but the same trend generalizes to majority groups) and the minority group. Specifically, we observe a \textit{large generalization gap for the minority group}, indicative of overfitting—a point that has not been sufficiently highlighted in prior research.

To investigate the underlying causes of this generalization failure in the minority group, we turn to an analysis of memorization. Indeed, deep neural networks, as high-capacity models, are capable of fitting complex and atypical examples, making it reasonable to associate this generalization failure with memorization. Leveraging recent advances in understanding memorization, we employ the method proposed by \citet{maini2023can} to detect memorization. For a given input, the technique identifies the minimal set of neurons (or channels in convolutional layers) that, when removed: (i) flip the prediction of the target input, and (ii) preserve predictions on a reference sample (batch) of data. They are found via a heuristic search among neurons of layers. Given a layer index $l$ and neuron index $j$, neurons are denoted $z^{(l,j)}$. In other words, the technique performs a greedy iterative search across layers to locate neurons that, when dropped: (i) enable flipping the considered example’s prediction, indicated by high cross-entropy loss on this example, while (ii) preserving predictions on a reference sample (batch) of data, reflected in low cross-entropy loss on this sample.
More formally, given an input $\vx_i$, this is technically done by sequentially finding the neurons, for each iteration (per layer)
\begin{align}\label{eq:pred}
    z^{(l^*,j^*)} \in \argmax_{l,j} \Biggl[& \nabla_{\vtheta_l}\left(\mathcal{L} \left(f_{\hat{\vtheta}}(\vx_i), y_i\right) \right. \nonumber  \\
    &\left. - \frac{1}{|\mathcal{B}|} \sum_{(\vx, y) \in \mathcal{B}} \mathcal{L}\left(f_{\hat{\vtheta}}(\vx), y \right)\right)
    \Biggl]_j,
\end{align}
where $\mathcal{B}$ is the random batch on which the predictions have to be preserved, $f_{\hat{\vtheta}}$ is the current iteration of the modified model (where a neuron was dropped in the previous iteration), $\vtheta_l$ are parameters of the layer indexed by $l$, and $[.]_j$ represents the coordinate $j$ of the vector. Note that, at each iteration per layer, once neurons have been selected according to Eq.~\ref{eq:pred}, they are dropped, leading to $f_{\hat{\vtheta}}$. Then, the sequential procedure continues until the prediction for $\vx_i$ is flipped. The final neurons $z^{(l^*,j^*)}$ are seen as the most critical neurons uniquely associated with the given example. \citet{maini2023can} demonstrate that the proportion of these neurons can be used to detect memorized examples.

\begin{figure}
\centering
\includegraphics[width=.5\textwidth]{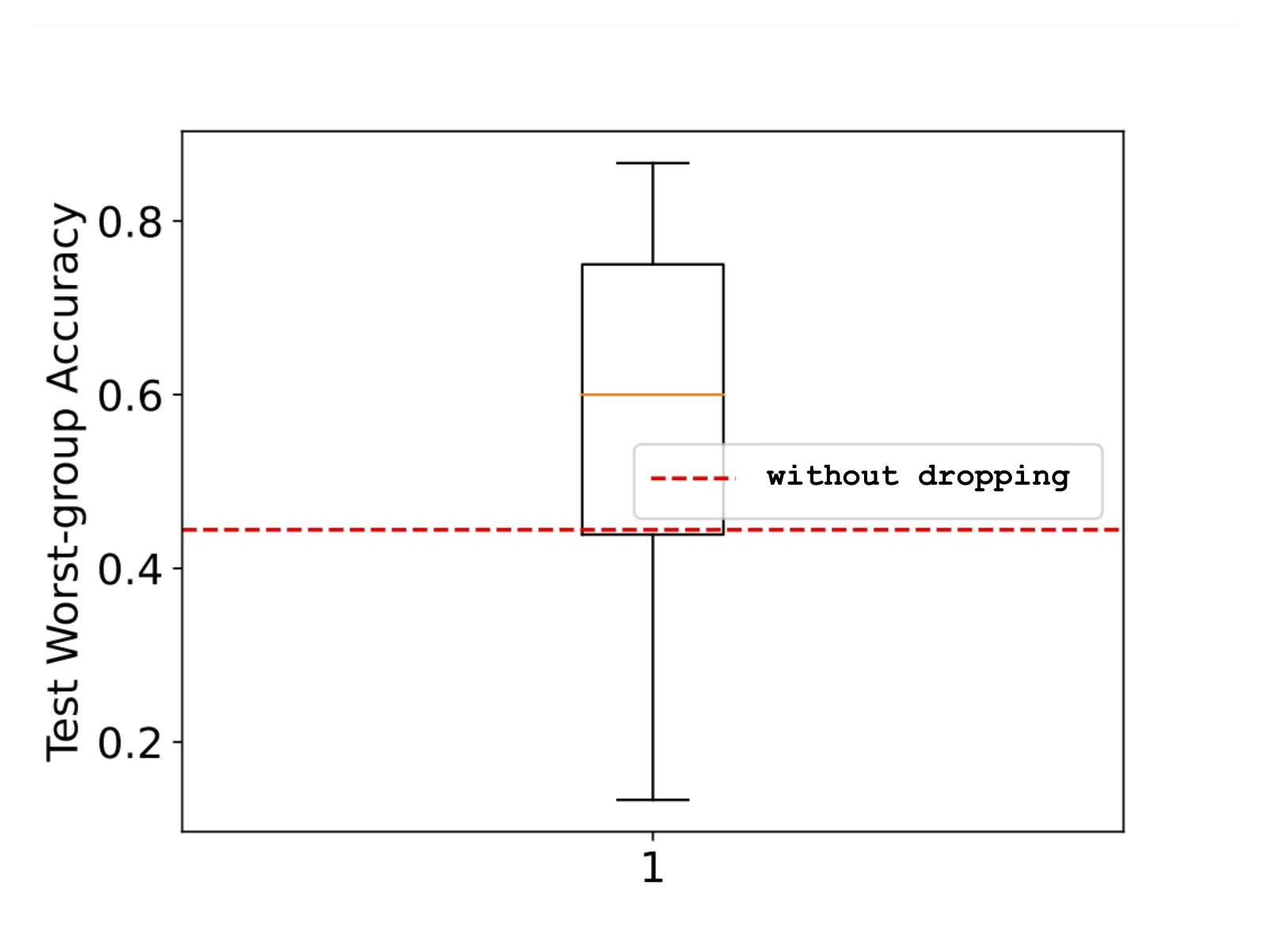}
\caption{Effect on the test worst-group accuracy when dropping memorizing neurons as shown in \ref{fig:analyze_memorization}. For each example in the minority-group sample, we drop their most critical neurons (memorizing neurons in this case), and report the measured test worst-group accuracy. From the quartiles on this figure, we observe that in $\approx 75\%$, of cases dropping out \textit{memorizing} neurons improves test worst-group accuracy.
  }
\label{fig:droping_out_neurons}
\end{figure}

\begin{figure*}[!t]
\centering
\includegraphics[width=.8\textwidth]{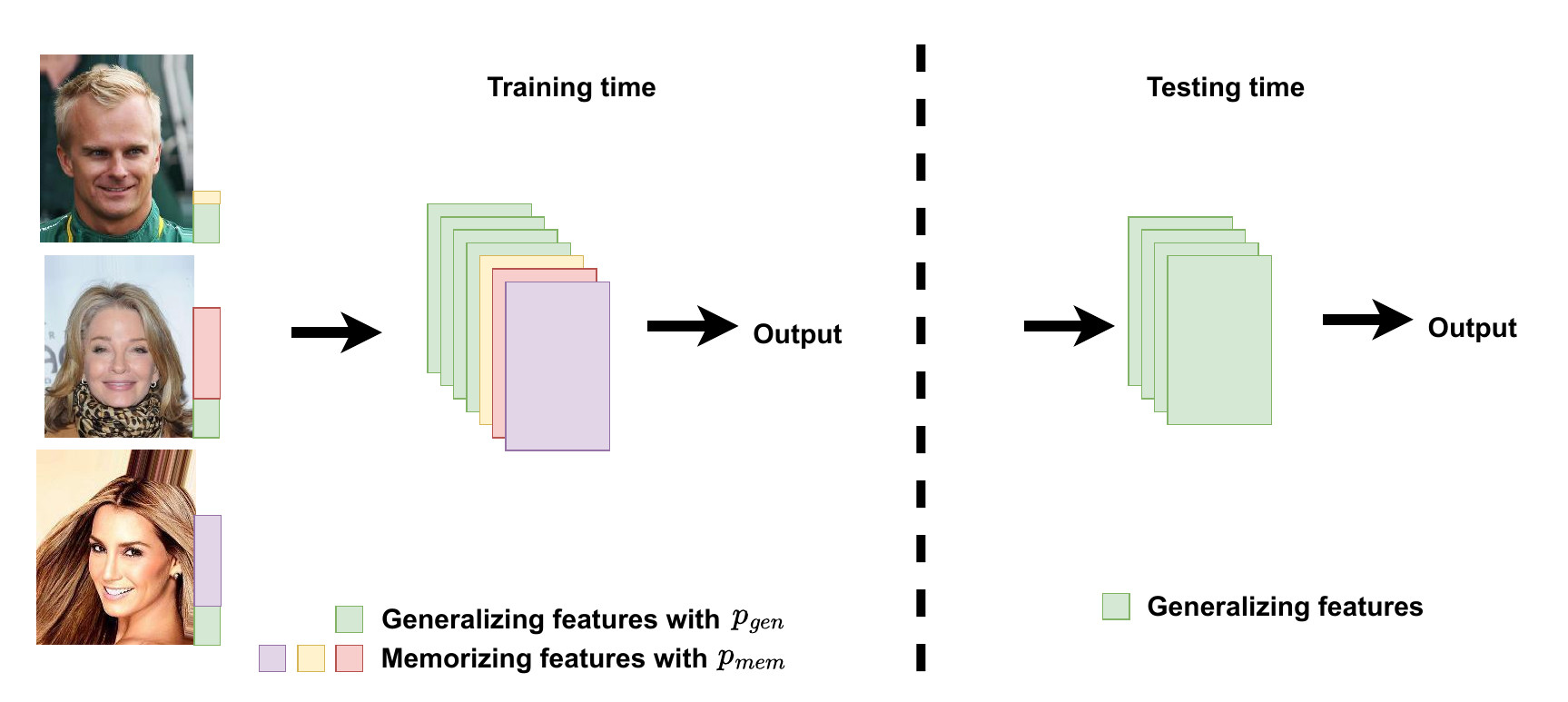}
\caption{Example-Tied Dropout as a FairDropout. The FairDropout redirects the example memorization on specific neurons. Memorizing neurons are uniformly allocated to training examples during training. During testing, these memorizing neurons are dropped.}
\label{fig:fairdropout}
\end{figure*}

We conduct this experiment to analyze the memorization behaviors in the context of spurious correlation on the CelebA dataset. Fig.~\ref{fig:analyze_memorization} shows the number of neurons required to flip the prediction of each example of a subsample of majority and minority groups. In general, we observe that, (i) the number of neurons required to flip each prediction from the minority group is considerably lower than the corresponding number for majority groups (see Fig.~\ref{fig:number_neurons}). Furthermore, as shown in Fig.~\ref{fig:impact_worst_group_accuracy}, (ii) these neurons have a smaller effect on training worst-group accuracy compared to those from the majority group. Referring to similar analyses in \citet{maini2023can}, (i) and (ii) indicate that minority group examples are more prone to memorization issues. 

While we have identified generalization problems and memorization issues within the minority group, there is no direct evidence suggesting a causal link between the two phenomena in the context of spurious correlation. To investigate this potential link, we conducted a new experiment to measure changes in test worst-group accuracy after dropping critical neurons of examples from the minority group. 

Fig.~\ref{fig:droping_out_neurons} shows the test worst-group accuracy  after individually dropping neurons for each example of the minority group.  We observe that for approximately $75\%$ of these drops, the worst-group accuracy significantly improves. This suggests that most of the neurons identified as memorizing neurons are detrimental to minority-group generalization.

\begin{figure*}[!t]
\centering
\begin{subfigure}[]{0.32\linewidth}
\includegraphics[width=\textwidth]{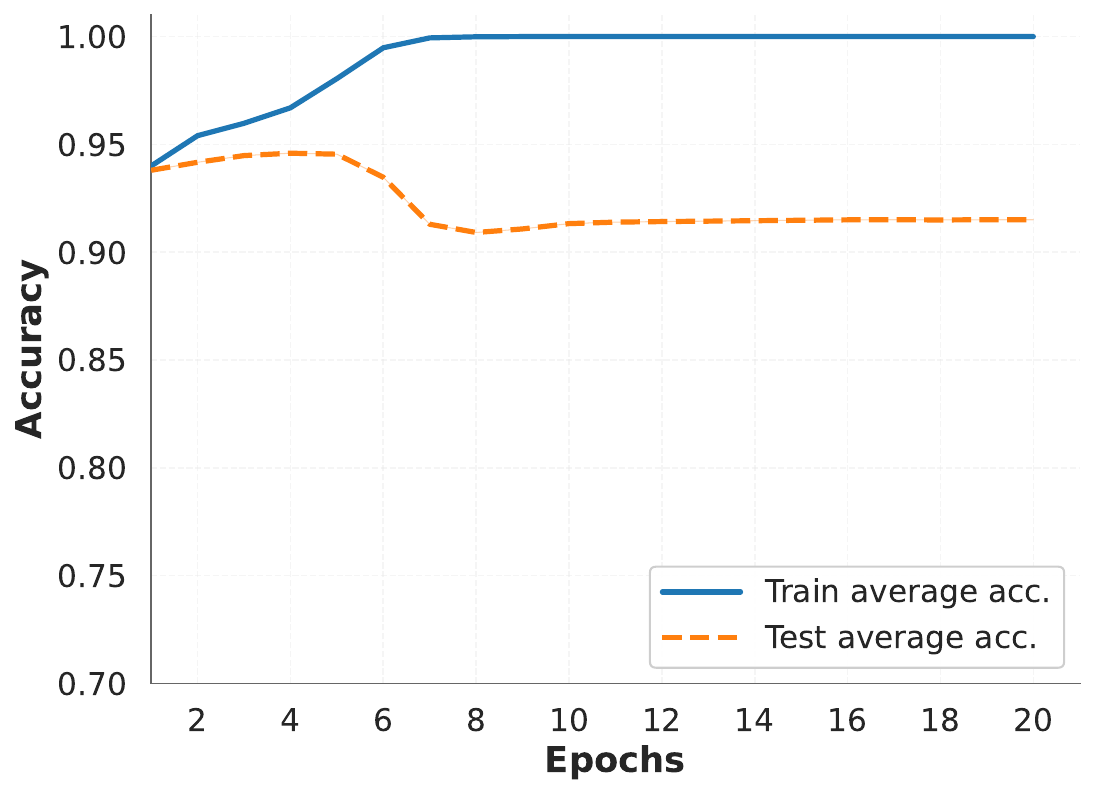}
\end{subfigure}
\begin{subfigure}[]{0.32\linewidth}
\includegraphics[width=\textwidth]{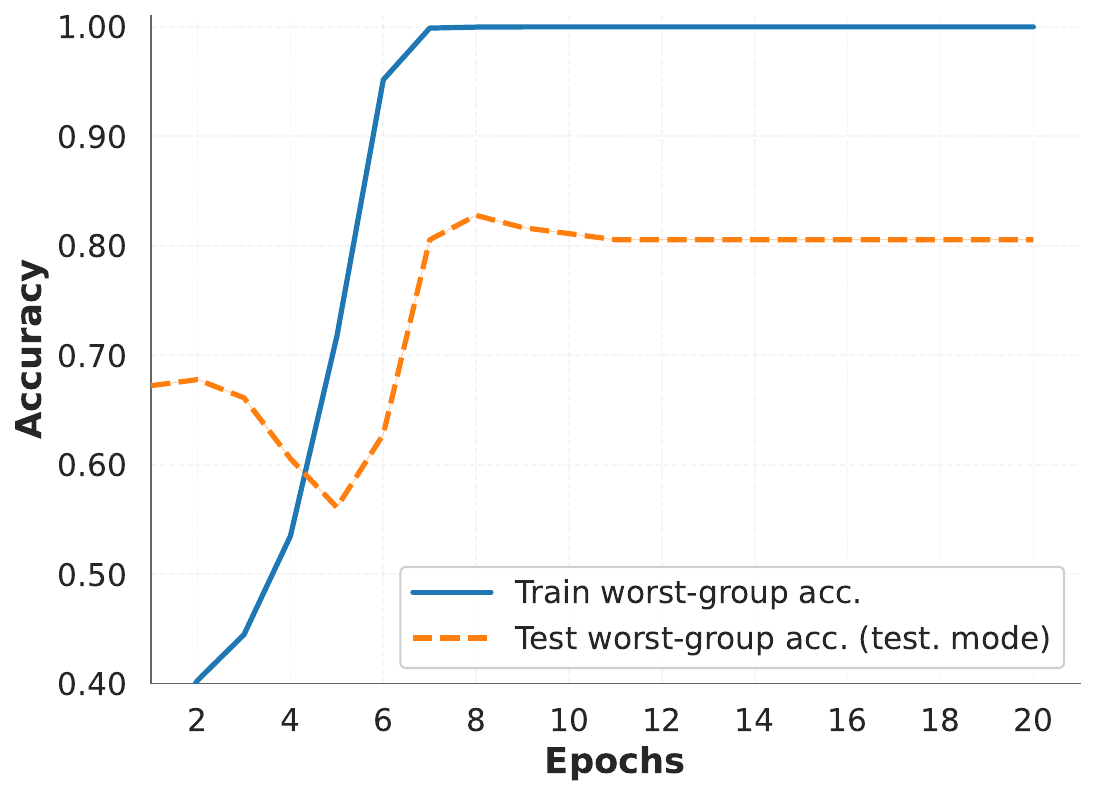}
\end{subfigure}
\begin{subfigure}[]{0.32\linewidth}
\includegraphics[width=\textwidth]{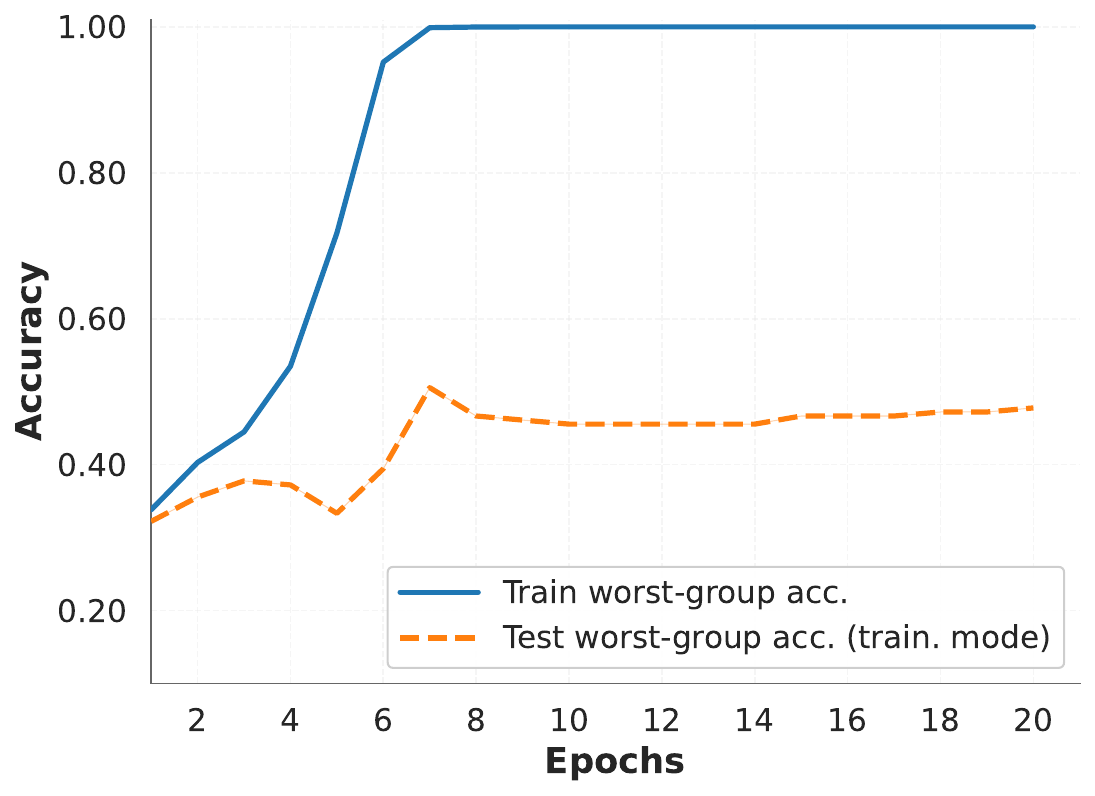}
\end{subfigure}
\caption{Training with FairDropout on CelebA. Left: train/test averages are plotted in the testing mode. Center and right: train/test worst-group accuracy with FairDropout are plotted. Training and testing mode respectively refer to the evaluation without dropping memorizing neurons, and after dropping them. We observe that dropping out these memorizing neurons has the benefit of improving worst-group accuracy.}
\label{fig:faidropout_effect}
\end{figure*}

\subsection{The Example-TiedDropout as a FairDropout}
After observing that certain neurons are closely tied to specific examples--particularly the memorizing ones in the minority group-- and that dropping these neurons positively impacts minority group generalization, it becomes crucial to leverage this insight by directing this memorization to fixed neurons. Drawing inspiration from the example-tied dropout introduced in the context of label noise by \citet{maini2023can} for relatively small networks (such as ResNet-9) and smaller datasets (like MNIST~\citep{deng2012mnist} or CIFAR-10~\citep{krizhevsky2009learning}), we introduce FairDropout, an example-tied dropout method specifically designed to address spurious correlation. Unlike the original example-tied dropout, FairDropout can be applied not only after any intermediate layers but also after newly added projection layers before the linear head.

As an example-tied dropout, FairDropout is a layer without learnable parameters, that divides neurons into two types, governed by two hyper-parameters: $p_\text{gen}$ and $p_\text{mem}$. The first set of neurons contains the generalizing neurons, which are seen by every example in the dataset. If the preceding layer of the FairDropout has the size $H$, then $p_\text{gen}H$ neurons are designated as generalizing. The remaining $(1-p_\text{gen})H$ neurons are considered as memorizing neurons and each example is allocated a set of memorizing neurons uniformly sampled with probability $p_\text{mem}$ from the $(1-p_\text{gen})H$ memorizing neurons. The \textit{fair} prefix comes from the fact that every example allocates the same fixed number of memorizing neurons. As depicted in Fig.~\ref{fig:fairdropout}, during training, given an example, and layer features as inputs of the FairDropout, the FairDropout outputs its generalizing features and its example-wise memorizing ones. In this case, each image allocates only one memorizing neuron. During testing, the memorizing ones are dropped. Finally, we observe that when $p_\text{gen}=1$ the FairDropout is just an identity function and trained models correspond to ERM-trained models.

\section{Experimental Results}
\label{sec:experiments}
We conduct experiments to evaluate the FairDropout on CelebA as a sanity check and on a benchmark suite.

\subsection{Warm-up on CelebA: FairDropout Balances Group Accuracy}
We applied the FairDropout after the third residual block on ResNet-50, with the hyperparameters $p_\text{mem} = p_\text{gen} = 0.2$, and monitor the train/test average/worst-group accuracy. 

Fig.~\ref{fig:faidropout_effect} shows the evolution of the train/test average and worst-group accuracy throughout epochs. It can be observed on the right-most plot that evaluating accuracy in FairDropout's training mode (where memorizing neurons are retained) yields results nearly identical to those without FairDropout, effectively mirroring ERM behavior: the worst-group accuracy plateaus around $45\%$, similar to Fig.~\ref{fig:minority_majority_groups}.

In contrast, in FairDropout's testing mode (where memorizing neurons are dropped), the worst-group accuracy does not plateau at $45\%$ but instead stabilizes around $80\%$. Thus, dropping memorizing neurons after training with FairDropout clearly boosts worst-group accuracy. In the next section, we compare FairDropout against state-of-the-art methods for handling spurious correlation.

\begin{table*}[!h]
\centering
\caption{Dataset overview with modality, number of attributes, classes, train, validatiand test set sizes, and group distributions.}
\label{tab:datasets}
\begin{tabular}{lcccccccc}
\toprule
\textbf{Dataset} & \textbf{Data} & \textbf{$|\mathcal{A}|$} & \textbf{$|\mathcal{Y}|$} & \textbf{$|\mathcal{D}_\text{tr}|$} & \textbf{$|\mathcal{D}_\text{val}|$}  & \textbf{$|\mathcal{D}_\text{test}|$}  & \textbf{Max group (\%)} & \textbf{Min group (\%)} \\
\midrule
Waterbirds & Image & 2 & 2 & 4795 & 1199 & 5794 & 3498 (73.0\%) & 56 (1.2\%) \\
CelebA & Image & 2 & 2 & 162770 & 19867 & 19962 & 71629 (44.0\%) & 1387 (0.9\%) \\
MetaShift & Image & 2 & 2 & 2276 & 349 & 874 & 789 (34.7\%) & 196 (8.6\%) \\

MultiNLI & Text & 2 & 3 & 206175 & 82462 & 123712 & 67376 (32.7\%) & 1521 (0.7\%) \\
MIMIC-CXR & X-rays & 6 & 2 & 303591 & 17859 & 35717 & 68575 (22.6\%) & 7846 (2.6\%)  \\
\bottomrule
\end{tabular}
\end{table*}
\subsection{Benchmarking FairDropout with Baselines}
Before presenting the results of the comparison between baselines, we present the experimental setup used largely inspired from \citet{yang2023change}.
\subsubsection{Experimental Setup}
\label{sec:setup}

We use the recently proposed subpopulation shift library and benchmark suite~\citep{yang2023change} that implements the state-of-the-art methods in spurious correlation.

We use 5 diverse datasets that are very used in spurious correlation litterature~\citep{yang2023change}. Characteristics of these datasets are summarized in Table~\ref{tab:datasets}.

\textbf{CelebA.} As introduced in Sec.~\ref{sec:problem_desc}, CelebA~\citep{liu2015deep} is one of the largest, real-world image datasets used in the context of spurious correlation. It has around 200,000 celebrity face images. The task, in the spurious correlations literature, is to predict the hair color of persons (blond vs. non-blond) and the spurious correlation is the gender. 

\textbf{MetaShift.} The dataset Metashift~\citep{liang2022metashift} that we use here is an image dataset that was built by \citet{yang2023change}. The goal is to distinguish between the two animals (cats vs dogs). The spurious attribute is the image background. Cats are more likely to be indoors, while
dogs are more likely to be outdoors.

\textbf{Waterbirds.} Waterbirds~\citep{wah2011caltech} is a well-known \textit{synthetic} image dataset for binary classification. The task is to classify whether a bird is a landbird or a waterbird. The spurious attribute is the background (water or land). There are therefore 4 groups that are from $\{\text{landbird}, \text{waterbird}\}\times\{\text{water background, land background}\}$. 

\textbf{MultiNLI.} The MultiNLI dataset~\citep{liang2022metashift} is a text dataset very used in spurious correlation literature. The target is the natural language relationship between the premise and the hypothesis. It has three classes (neutral, contradiction, or entailment). The spurious attribute is a variable that tells whether negation appears in the text or not. Indeed, negation is highly correlated with the contradiction label.

\textbf{MIMIC-CXR.} MIMIC-CXR~\citep{johnson2019mimic} is a chest X-ray dataset, where its approximately 300,000 images come from the Beth Israel Deaconess
Medical Center from Boston, Massachusetts. We use the setting of \citet{yang2023change}, where the label is “No Finding” as the label. The positive class means that the patient is not ill. the spurious attribute domain is the cross-product of race (White, Black, Other) and gender.

All the data preprocessing and train/val/test splits are directly adopted from~\citet{yang2023change} as we implement our method in their library. 

\textbf{Models.} As in the benchmark~\citep{yang2023change}, we use the Pytorch pretrained ResNet-50 models for image datasets and BERT~\citep{sung2019pre} for the MultiNLI text datasets.

\textbf{Metrics.} According to most previous works, we evaluate the reliance on spurious correlation through worst-group accuracy.

\begin{table*}[t]
\caption{Comparison of the FairDropout against state-of-the-art methods when spurious attribute annotations or group annotations are unknown in both train and validation sets. Test worst-group accuracy is reported and is obtained from the subpopulation shift benchmark~\citep{yang2023change}. 
Bold values indicate the highest accuracy achieved, while underlined values represent the second-best accuracy in the dataset column.
}
\label{tab:results}
\centering
\resizebox{\textwidth}{!}{%
\begin{tabular}{@{}ll cc cc cc cc cc@{}}
\toprule
\textbf{Category} & \textbf{Method} & \textbf{CelebA} & \textbf{MetaShift} & \textbf{Waterbirds} & \textbf{MultiNLI} & \textbf{MIMIC-CXR} \\
\midrule
Standard 
    & ERM 
    & 69.1 $\pm$ 4.7 & 82.1 $\pm$ 0.8 & 57.6 $\pm$ 0.8 & 66.4 $\pm$ 2.3 & 68.6 $\pm$ 0.2 \\

\cmidrule(r){1-2}
Data Augmentation 
    & Mixup 
    & 57.8 $\pm$ 0.8 & 79.0 $\pm$ 0.8 & 77.5 $\pm$ 0.7 & 66.8 $\pm$ 0.3 & 66.8 $\pm$ 0.6 \\

\cmidrule(r){1-2}
\multirow{6}{*}{Spurious Correlation}
    & \textbf{FairDropout (ours)} 
    & \underline{75.6} $\pm$ 2.1 & \underline{85.9} $\pm$ 1.1 & 77.8 $\pm$ 0.5 & \underline{70.3} $\pm$ 2.4 & \textbf{70.6} $\pm$ 0.6 \\
    
    & GroupDRO
    & 68.3 $\pm$ 0.9 & 83.1 $\pm$ 0.7 & 73.1 $\pm$ 0.4 & 64.1 $\pm$ 0.8 & 67.4 $\pm$ 0.5 \\
    
    & CVaRDRO
    & 60.2 $\pm$ 3.0 & 83.5 $\pm$ 0.5 & 75.5 $\pm$ 2.2 & 48.2 $\pm$ 3.4 & 68.0 $\pm$ 0.2 \\
    
    & JTT 
    & 48.3 $\pm$ 1.5 & 82.6 $\pm$ 0.4 & 71.2 $\pm$ 0.5 & 65.1 $\pm$ 1.6 & 64.9 $\pm$ 0.3 \\
    
    & LfF 
    & 53.0 $\pm$ 4.3 & 72.3 $\pm$ 1.3 & 75.0 $\pm$ 0.7 & 57.3 $\pm$ 5.7 & 62.2 $\pm$ 2.4 \\
    
    & LISA
    & 57.8 $\pm$ 0.8 & 79.0 $\pm$ 0.8 & 77.5 $\pm$ 0.7 & 66.8 $\pm$ 0.3 & 66.8 $\pm$ 0.6 \\

\cmidrule(r){1-2}
\multirow{7}{*}{Imbalanced Learning}
    & ReSample 
    & 74.1 $\pm$ 2.2 & 81.0 $\pm$ 1.7 & 70.0 $\pm$ 1.0 & 66.8 $\pm$ 0.5 & 67.5 $\pm$ 0.3 \\
    
    & ReWeight 
    & 69.6 $\pm$ 0.2 & 83.1 $\pm$ 0.7 & 71.9 $\pm$ 0.6 & 64.2 $\pm$ 1.9 & 67.0 $\pm$ 0.4 \\
    
    & SqrtReWeight 
    & 66.9 $\pm$ 2.2 & 82.6 $\pm$ 0.4 & 71.0 $\pm$ 1.4 & 63.8 $\pm$ 2.4 & 68.0 $\pm$ 0.4 \\
    
    & CBLoss 
    & 65.4 $\pm$ 1.4 & 83.1 $\pm$ 0.0 & 74.4 $\pm$ 1.2 & 63.6 $\pm$ 2.4 & 67.6 $\pm$ 0.3 \\
    
    & Focal 
    & 56.9 $\pm$ 3.4 & 81.0 $\pm$ 0.4 & 71.6 $\pm$ 0.8 & 62.4 $\pm$ 2.0 & 68.7 $\pm$ 0.4 \\
    
    & LDAM 
    & 57.0 $\pm$ 4.1 & 83.6 $\pm$ 0.4 & 70.9 $\pm$ 1.7 & 65.5 $\pm$ 0.8 & 66.6 $\pm$ 0.6 \\
    
    & BSoftmax 
    & 69.6 $\pm$ 1.2 & 82.6 $\pm$ 0.4 & 74.1 $\pm$ 0.9 & 63.6 $\pm$ 2.4 & 67.6 $\pm$ 0.6 \\

\cmidrule(r){1-2}
Classifier Retraining
    & DFR
    & 73.7 $\pm$ 0.8 & 81.4 $\pm$ 0.1 & \underline{89.0} $\pm$ 0.2 & 63.8 $\pm$ 0.0 & 67.1 $\pm$ 0.4 \\
    
    & CRT 
    & 69.6 $\pm$ 0.7 & 83.1 $\pm$ 0.0 & 76.3 $\pm$ 0.8 & 65.4 $\pm$ 0.2 & 68.1 $\pm$ 0.1 \\
    
    & ReWeightCRT 
    & 70.7 $\pm$ 0.6 & 85.1 $\pm$ 0.4 & 76.3 $\pm$ 0.2 & 65.2 $\pm$ 0.2 & 67.9 $\pm$ 0.1 \\
        & \textbf{FairDropout-DFR (ours)}
    & \textbf{77.6} $\pm$ 1.1 &  \textbf{87.7} $\pm$ 0.7  &  \textbf{89.9} $\pm$ 0.3  & \textbf{71.8} $\pm$ 1.1  & \underline{70.3} $\pm$  0.6\\
\bottomrule
\end{tabular}
}
\vspace{0.5em}
\raggedright
\end{table*}

\textbf{Baseline methods.} We compare the FairDropout with state-of-the-art algorithms implemented in the subpopulation shift benchmark. Our work does not need the knowledge of group information. We thus evaluate our method in the setting where we do not have group information. However, methods that need group information have been converted by \citet{yang2023change} to an equivalent method by considering class information instead of group. For example, GroupDRO can be converted by an equivalent goal of minimizing worst-class error. Benchmarked methods from the spurious correlation literature  include GroupDRO~\citep{sagawa2019distributionally}, CVaRDRO~\citep{duchi2021learning}, JTT~\citep{liu2021just}, LfF~\citep{nam2020learning}, LISA~\citep{yao2022improving}. There are also two-phase methods that retrain the classifier, which are DFR~\citep{yao2022improving} (retraining is done on the \textit{validation} set), CRT and its variant ReWeightCRT~\citep{kangdecoupling}. Finally, we also include methods that are mostly designed for the imbalanced learning problem, which are ReSample~\citep{japkowicz2000class}, ReWeight~\citep{japkowicz2000class}, SqrtReWeight~\citep{japkowicz2000class}, CBLoss~\citep{cui2019class}, LDAM~\citep{cao2019learning} and BSoftmax~\citep{ren2020balanced}. Note that the FairDropout technique can be combined with any of these baseline methods to boost its performance. For instance, we have applied DFR after training models with FairDropout, leading to \textit{FairDropout-DFR};

\textbf{Hyperparameter tuning.} As we consider the most difficult case we do not have group information for the training and validation sets, similarly with the benchmark \citet{yang2023change}, we tune the $p_\text{mem}, p_\text{gen}$, learning rate, and weight decay with the worst-class accuracy. We use the SGD optimizer with weight decay.

\textbf{Positions of the FairDropout Layers.}
In principle, the FairDropout layer can be placed after any intermediate layer in the network. However, in large-scale, potentially pre-trained models, the placement of FairDropout may require careful consideration. In models with skipped connections as in ResNet-50, in our settings, we consider possible positions after residual blocks. In BERT-like models, we propose adding a new linear layer before the classifier head and positioning the FairDropout layer there. This ensures that the pre-trained features are preserved while FairDropout controls memorization during fine-tuning. The optimal placement, however, depends on the dataset, as spurious correlations exhibit task-specific levels of abstraction. Therefore, we tune the position of FairDropout along with other optimization hyperparameters using worst-class accuracy as a guiding metric.

\subsubsection{Results and Discussion}
We report the worst-group accuracy results obtained after running our FairDropout method on the subpopulation shift library, averaged over 3 independent runs. Table~\ref{tab:results} presents these results with methods categorized according to the presentation done in Sec.~\ref{sec:setup}, following \citet{yang2023change}. As a reminder, in this setting, the spurious attribute and the group annotations are \textbf{unavailable in both the training and validation datasets}. All the methods or their adapted version are tuned with worst-class accuracy and the results come from \citet{yang2023change}.

The table allows us to make the following observations. Our FairDropout method consistently outperforms ERM by a large margin across all datasets. 

We can also observe that FairDropout outperforms or has comparable performance to state-of-the-art methods for handling spurious correlation (e.g., GroupDRO, JTT) and class imbalance (e.g., ReSample, Focal Loss) across all benchmarks. More specifically, the datasets on which FairDropout achieves the most successful results are MultiNLI ($70.3 \pm 2.4$) and MIMIC-CXR ($70.6\pm 0.6$). 

On CelebA and MetaShift, while FairDropout outperforms its competitors from the same family (spurious correlation methods), its performance is comparable to Resample on CelebA ($75.6\pm2.1$ vs $74.1\pm 2.2$) and ReWeightCRT on Metashift ($85.9\pm 1.1$ vs $85.1\pm 0.4$). It is worth mentioning that our models with the FairDropout are trained with classic cross-entropy, meaning its performance \textbf{can be further enhanced} with any of these existing imbalance-aware learning methods. 

On the Waterbirds dataset, while FairDropout improves upon ERM, its performance is below (or closely matches) some classifier retraining methods, eg., DFR. Since Waterbirds is a dataset \textit{synthetically} generated by placing bird objects into different backgrounds, it has already been observed that ImageNet pre-trained ImageNet features can be effectively transferred~\citep{izmailov2022feature} without finetuning the entire model, which may explain the superior performance of DFR. We further propose to consider applying DFR after training models with FairDropout. This is therefore done with a classifier retraining on a balance-class validation set. We call this method \textit{FairDropout-DFR}.

The integration of FairDropout with DFR (FairDropout-DFR) achieves state-of-the-art performance, outperforming all standalone methods on four of five datasets. For instance, on CelebA, FairDropout-DFR attains $77.6 \pm 1.1$, a $3.9\%$ improvement over DFR alone ($73.7\pm0.8$), demonstrating that reducing memorization and spurious correlations during training can amplify classifier retraining benefits. Similarly, on Waterbirds, FairDropout-DFR ($89.9\pm0.3$) slightly exceeds even the strong performance of DFR ($89.0\pm0.2$), resolving the apparent gap observed with standalone FairDropout. This suggests that FairDropout and DFR are complementary: while DFR effectively leverages pre-trained representations, FairDropout can benefit from it when combined.

Overall, the results obtained demonstrate the effectiveness of FairDropout in reducing reliance on spurious correlations without requiring explicit group labels. FairDropout excels on natural distribution shifts (e.g., MultiNLI, MIMIC-CXR), but its gains are less pronounced on synthetic benchmarks like Waterbirds. Crucially, FairDropout synergizes well with classifier retraining techniques, pushing performance boundaries in challenging settings.



\section{Limitations and Conclusion}

In this paper, we explored, for the first time, the lack of generalization of minority-group examples and its connection to memorization in the context of spurious correlations. To address this, we introduced FairDropout, an example-tied dropout technique that can be applied for large pre-trained networks, aimed at reducing reliance on spurious features.

FairDropout enables the localization of memorization by directing spurious features to specific neurons that, when dropped during inference, can enhance worst-group accuracy. Through empirical evaluation, we demonstrated that FairDropout outperforms several baseline methods across diverse datasets spanning image, medical, and language tasks.

Nonetheless, our study has certain limitations. First, prior research has shown that memorization can sometimes be beneficial for generalization~\citep{feldman2020does}, an aspect that warrants further investigation, particularly for certain applications and datasets. Second, while FairDropout implicitly assumes that \textit{generalizing} neurons are less likely to memorize examples—since memorization is predominantly managed by the designated memorizing neurons—this hypothesis requires further exploration, which lies beyond the scope of this paper. 


\newpage
\nocite{langley00}

\bibliography{example_paper}
\bibliographystyle{icml2025}

\newpage
\appendix
\onecolumn

\clearpage
\setcounter{page}{1}
\section{Appendix}

\subsection{Hyperparameters}
Table~\ref{tab:hyper_param_range} describes the range of hyperparameters that we used to tune the hyperparameters.
\begin{table}[h]
\caption{Hyperparameter ranges. Here dp$i$ stands for the position before the residual layer block $i$, dplogits stands for the position before the linear classifier head, dpfc stands for a position before the classifier head but after a newly introduced linear projection layer.}
\label{tab:hyper_param_range}
\centering
\resizebox{0.5\textwidth}{!}{%
\begin{tabular}{c|c}
Hyperparameters & sets\\
\hline
learning rate & $\{1e-3, 1e-4, 1e-5\}$\\
Weight decay & $\{1e-3, 1e-4, 1e-5, 1e-6\}$\\
$p_\text{gen}$ & $\{.2, .3, .4, .5, 6\}$ \\
$p_\text{mem}$ & $\{.001, .1, .2, .4\}$ \\
FairDropout positions on ResNet-50 & \{dp2, dp3, dp4, dp5 \} \\
FairDropout positions on BERT & \{dplogits , dpfc\}
\end{tabular}%
}
\end{table}

\subsection{Hyperparameter Sensitivity}
Table~\ref{tab:sensitivity} shows the sensitivity of $p_\text{gen}$ under low $p_\text{mem}$ on two datasets. One can observe that under low $p_\text{mem}$ does not need to be high.

\begin{table}[h]
\caption{Sensitivity of $p_\text{gen}$ under low $p_\text{mem}$.}
\label{tab:sensitivity}
\centering
\begin{tabular}{c|cc|c}
\toprule
Datasets & $p_\text{gen}$ & $p_\text{mem}$ & Worst-group accuracy \\
\midrule
\multirow{3}{*}{CelebA} & 0.5 & 0.1 & 56.14 \\
& 0.4 & 0.1  &  59.54\\
& 0.2 & 0.1 & 71.66\\
\midrule
\multirow{3}{*}{MIMIC-CXR} & 0.5 & 0.1 & 63.15\\
 & 0.4 & 0.1 & 65.43 \\
 & 0.2 & 0.1 & 66.20\\
\bottomrule
\end{tabular}%
\end{table}







\end{document}